\documentclass[twoside,12pt]{article}

\usepackage{blindtext}
\usepackage[T1]{fontenc}
\usepackage[utf8]{inputenc}
\usepackage{microtype}
\usepackage[english]{babel}
\usepackage{floatpag}

\usepackage[hmarginratio=1:1,top=32mm,columnsep=20pt,tmargin=1.1in, lmargin=1in, rmargin=1in, bmargin=1.0in]{geometry}
\usepackage[hang, small,up,up]{caption}%,textfont=it,labelfont=bf
\usepackage{booktabs}
\usepackage{enumitem}
\setlist[itemize]{noitemsep}

\usepackage[symbol]{footmisc}

\usepackage{natbib}
\bibpunct{(}{)}{;}{a}{}{;}

\usepackage{graphicx}
\usepackage{amsmath}
\usepackage{amssymb}
\usepackage{graphicx}
\usepackage{wrapfig}
\usepackage{xcolor}
\usepackage{subcaption}
\usepackage[ruled,vlined]{algorithm2e}
\usepackage[symbol]{footmisc}
\usepackage[colorinlistoftodos,prependcaption]{todonotes}

\usepackage{abstract}

\usepackage{fancyhdr}
\usepackage{titling}
\usepackage{hyperref}

\DeclareMathOperator*{\argmax}{arg\,max}

\newcommand{\C}{\mathcal{C}}

\pretitle{\begin{center}\Huge\bfseries} % Article title formatting
\posttitle{\end{center}} % Article title closing formatting
\title{\Large On Hyperparameter Search \\in Cluster Ensembles\\} % Article title
\author{%
\normalsize \textsc{Luzie Helfmann}\footnotemark[1]\\%\thanks{The authors contributed equally.} \\
\normalsize \textsc{Johannes von Lindheim}\footnotemark[1]
\\[1ex]
\normalsize Zuse-Institute Berlin \& \\[-0.18cm]
\normalsize Freie Universität Berlin \\[-0.18cm]
\normalsize Institut für Mathematik\\
\normalsize helfmann@zib.de\\[-0.18cm]
\normalsize lindheim@zib.de
\and
\normalsize \textsc{Mattes Mollenhauer}\footnotemark[1] \\
\normalsize \textsc{Ralf Banisch}
\\[1ex]
\normalsize Freie Universität Berlin \\[-0.18cm]
\normalsize Institut für Mathematik\\
\normalsize mattes.mollenhauer@fu-berlin.de\\[-0.18cm]
\normalsize ralf.banisch@fu-berlin.de\\
}
\date{}
\renewcommand{\maketitlehookd}{%
\begin{abstract}
\noindent %
Quality assessments of models in unsupervised learning and clustering
verification in particular
have been a long-standing problem in the machine learning research.
The lack of robust and universally applicable cluster validity scores often makes the
algorithm selection and hyperparameter evaluation a tough guess.
In this paper, we show that cluster ensemble aggregation techniques such as consensus
clustering
may be used to evaluate clusterings and their hyperparameter configurations.
We use normalized mutual information
to compare individual objects of a clustering
ensemble to the constructed consensus of the whole ensemble and show, that the resulting score
can serve as an
overall quality measure for clustering problems. This method
is capable of highlighting the standout clustering and hyperparameter configuration in the
ensemble even in the case of a
distorted consensus.
We apply this very general framework to various data sets and give possible directions for
future research.
\end{abstract}
\footnotetext[1]{The authors contributed equally.}
}
\begin{document}

% Print the title
\maketitle

%%%%%%%%%%%%%%%%%%%%%%%%%%%%%%%%%%%%%%%%%%%%%%%%%%%%%%%%%

\section{Introduction}
Hyperparameter optimization for supervised learning models is classically accomplished
with respect to an external accuracy measure or
objective function defined on behalf of either class labels or regression values.
Different models and outcomes of different hyperparameter configurations can therefore be
scored and compared easily. The actual hyperparameter optimization
methods then depend on the chosen model and range from
random search methods \citep[see for example][]{bengio12} over brute force grid evaluation to
gradient descent variants and greedy methods on the
hyperparameter space \citep{kingma14}. However, the situation in unsupervised learning tasks
such as
data clustering is generally different
due to the lack of a universal interpretation of the validity of a model.
Over the years, various approaches to define clustering quality measures
have come up.

Internal measures such as the silhouette coefficient
\citep[see][]{rousseeuw87silhouette} or the
Dunn index \citep[see][]{dunn74}
make use of intrinsic properties of the data and the corresponding clustering.
This yields the fundamental drawback, that there is generally no internal measure
that applies for every type of problem or every clustering algorithm.
For example, an internal measure that is purely relying on inter- and intra-cluster distances
might be misleading
for data containing several accumulations of observations in elongated and nonconvex
geometrical shapes.
As another example, the notion of a cluster centroid might not be meaningful in scenarios with
clusters embedded
in highly nonlinear submanifolds.

External clustering quality measures like mutual information and its variants
\citep{vinh2010information} or the Rand index \cite[see][]{rand71}
are often derived by modifying already existing concepts from
information theory and statistics.
They rely on data from additional sources of knowledge. The additional information is mostly
given in the form of some external observation labeling or clustering benchmark on the same
data.
For classical pattern recognition tasks in unsupervised learning, labeled data will generally
not exist.
Therefore, one might use external measures to compare distinct clusterings in a pairwise
manner, yielding only a relative score.
This lack of a universal benchmark makes the search for a suitable clustering algorithm
and the optimal choice of a hyperparameter configuration extremely hard.

\begin{figure}[htb!]
\centering
\includegraphics[width=1\textwidth]{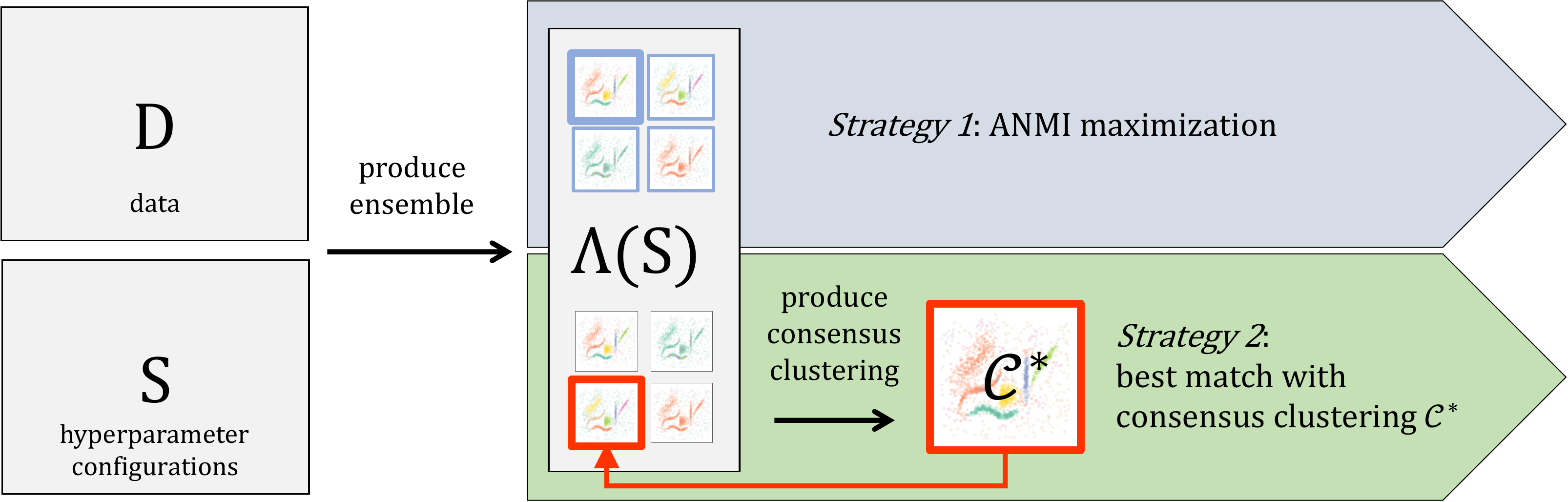}
\caption{Schema for our proposed method for finding clustering hyperparameters.
Here, $\Lambda (S)$ is an ensemble of clusterings of a data set $D$
based on a hyperparameter set $S$ and $\mathcal{C}^*$ is the consensus clustering of
$\Lambda(S)$.
For detailed explanations of the notation, see Section \ref{sec:cons} and Section
\ref{sec:hyperparam_search}.}
\label{fig: schemata}
\end{figure}

\noindent However, one may use
outcomes of clustering ensemble aggregation methods such as
consensus clustering as formulated by \citet{strehl2002cluster} as some type of constrained
ground truth labeling. In this paper, we propose to apply consensus clustering combined with
normalized mutual information \citep{strehl2002cluster, cover2006elements}
as an exemplary external validity measure to classical problems to perform cluster evaluation
and compare cluster algorithms and hyperparameters
respectively. If the
produced clustering ensemble for a specific data set is big and diverse enough,
evaluating the individual clusterings relatively to the
the aggregated clustering yields information about the quality of each clustering, see Figure
\ref{fig: schemata}. Furthermore, as a second strategy for validating individual clusterings,
we propose to compare them more directly with the clustering ensemble, in our case using the
average normalized mutual information.

This paper is structured as follows: In Section \ref{sec:background},
we give a concise overview of normalized mutual information
and consensus strategies for clustering problems and introduce our notation. In Section
\ref{sec:hyperparam_search},
we explain how these techniques can be combined to search for clustering hyperparameters.
We present a choice of numerical experiments in
Section \ref{sec:results}. Concluding with Section \ref{sec:conclusion}, we again highlight
the general features of our approach and give possible future directions.
Appendix A serves as an overview of classically used clustering validation metrics
which are employed for comparison in our experiments.

\section{Background on NMI and Consensus Clustering}
\label{sec:background}
We first introduce our notation and explain the different approaches behind several
classical clustering algorithms in Section \ref{sec:clustering}. We will illustrate, how the
shared information between two clusterings can be measured using the notion of normalized
mutual information (NMI) in Section \ref{sec:nmi}. Moreover, we explain how multiple
clusterings of the same data set can be combined into a so-called \emph{consensus clustering},
see Section \ref{sec:cons} and \ref{sec:consensuscompu}. This gives the
foundation of our proposed methods for finding hyperparameters in Section
\ref{sec:hyperparam_search}.

\subsection{The Clustering Problem}\label{sec:clustering}
The typical clustering problem setting is as follows:
A set $D=\{x_1, \dots, x_n\}\subseteq \mathbb{R}^d$ of data points is given.
The goal is to define a clustering $\mathcal{C} = \{C_1, \dots,C_k\}$ of $k$ clusters, such
that one is either able to map the $x_i$ directly onto the clusters $C_j$ (hard clustering),
or one maps the probability
vectors in the $k$-dimensional probability simplex onto the clusters $C_j$ (fuzzy clustering).
We consider the case of hard clustering in this article, so that $C_j\subset D, j=1, \dots, k$. Formally, $\mathcal{C} = \{C_1,
\dots,C_k\}$ is a hard clustering of $D$ if
\[
C_i \cap C_j = \emptyset \quad \mbox{and}\quad \bigcup_{i=1}^k C_i = D.
\]
The typical way of expressing a clustering is simply by labeling each
data point with the index of its corresponding cluster.
\subsection{Cluster comparison criteria}\label{sec:comparison}

Suppose two clusterings $\C$ and $\C'$ are given for the same problem. How should one measure
their similarity? In this section, we briefly review some comparison criteria available in the
literature.

Since the clusterings are often stored with lists of labels (where each data point is assigned a label or index of some cluster), a common requirement for any such similarity criterion is relabelling invariance, so that two clusterings $\C$ and $\C'$ with label lists $(1, 1, 2, 2, 2, 3)$ and $(3, 3, 1, 1, 1, 2)$ are considered equal. Viewed differently, permuting the indices of the clusters should not change the similarity measure.

A useful tool is the \emph{contingency table} or \emph{confusion matrix}. The confusion matrix for two
clusterings $\C = \{C_1,\ldots C_K\}$ and $\C' = \{C'_1, \ldots, C'_{K'}\}$ is a $K \times
K'$ matrix whose $kk'$th entry measures the amount of overlap between clusters $C_k$ and
$C'_{k'}$, i.e.
\[
n_{kk'} = | C_k \cup C'_{k'}|.
\]

One category of criteria is based on counting pairs of points. Let $N_{11}$ be the number of
pairs of points from $D$ that are in the same cluster under both $\C$ and $\C'$, $N_{00}$ the
number of pairs of points that are in different clusters under both $\C$ and $\C'$, $N_{10}$
the number of point pairs that are in the same cluster under $\C$ but different clusters under
$\C'$ and $N_{01}$ the number of point pairs that are in different clusters under $\C$ but
the same cluster under $\C'$. Further, let $n = |D|$. Popular criteria are the Rand index
$R(\C, \C')$ \citep{rand71} and Jaccard index $J(\C, \C')$ \citep{ben2001stability}
\[
R(\C, \C') = \frac{N_{00} + N_{11}}{n(n-1)/2}, \qquad J(\C, \C') = \frac{N_{11}}{N_{11} +
N_{01} + N_{10}}.
\]
In order to obtain an index that has range $[0,1]$, the adjusted Rand index $ARI(\C, \C')$ has
been introduced in \cite{rand71} and \cite{hubert85}
\[
ARI(\C,\C') = \frac{R(\C, \C') - \mathbb{E}[R]}{1-\mathbb{E}[R]},
\]
where $\mathbb{E}[R]$ is the expectation value of $R$ under a null model.
For the detailed definitions of
the measures used in our numerical experiments, see Appendix \ref{sec:appendix}.

A second category of criteria is based on set matching; each cluster in $\C$ is given a best
match in $\C'$ and then the total amount of `unmatched' probability mass is computed. See
\cite{meilua2001experimental} and \cite{larsen1999fast} for examples in this category. The
common problem of best matching criteria is that they ignore what happens to the unmatched
part of the clusterings.

A third category of criteria, which includes the one we use mainly in this paper, is based on
information theory. The common idea is to interpret a clustering $\C$ as a discrete valued
random variable representing the outcome of drawing a point in $D$ uniformly at random and
examining its label. Example criteria include the mutual information discussed below, various
normalized variants, and the variation of information considered in \cite{Meila2007}.

\subsection{Normalized Mutual Information}\label{sec:nmi}
For two discrete-valued random variables $X$ and $Y$, whose joint probability distribution is
$p_{X, Y}$, the mutual information $I(X, Y)$ is defined as
\[
I(X, Y) = \mathbb{E}_{p_{X,Y}}\left[ \log \frac{p_{X,Y}}{p_X p_Y} \right]
\]
where $\mathbb{E}_p$ denotes the expected value over the distribution $p$. The mutual
information measures how much information about $X$ is contained in $Y$ and vice versa.
$I(X,Y)$ is not bounded from above. In order to arrive at a comparison criterion with range
$[0,1]$, we define the normalized mutual information following \citep{strehl2002cluster}
\begin{equation}
\label{eq:NMI}
NMI(X, Y) = \frac{I(X, Y)}{\sqrt{H(X)H(Y)}}
\end{equation}
where
\[
H(X) = -\mathbb{E}_{p_X}\left[ \log p_X \right]
\]
is the entropy of $X$. The fact that $NMI(X,Y) \in [0,1]$ easily follows from the observation
that $I(X,Y) \leq \min(H(X), H(Y))$. Alternative normalizations are possible, e.g. by the
joint entropy $H(X,Y)$ \citet{yao2003information}, by $\max(H(X), H(Y))$ or by $\frac{1}{2}\
(H(X)+H(Y))$ \citet{kvalseth1987entropy}. We use the normalization \eqref{eq:NMI} in order to
be consistent with the framework in \cite{strehl2002cluster} and because of the similarity to
the Pearson correlation coefficient, which normalizes the covariance of two random variables.

The NMI score \eqref{eq:NMI} yields a similarity measure $\phi^{\text{NMI}}(\mathcal{C}_i,
\mathcal{C}_j)$ for the two clusterings $\C_i$ and $\C_j$ upon interpreting them as random
variables \cite[see][]{strehl2002cluster}. Let $n_s^i$ respectively $n_t^j$ denote the number
of $s$-labels in clustering $i$ respectively $t$-labels in clustering $j$. Moreover, let
$n_{s, t}$ denote the number of data points, that have label $s$ in clustering $i$ and label
$t$ in clustering $j$. Then we define
\[
\phi^{\text{NMI}}(\mathcal{C}_i, \mathcal{C}_j) =
%\frac{I(\mathcal{C}_i, \mathcal{C}_j)}{\sqrt{H(\mathcal{C}_i)H(\mathcal{C}_j)}} =
\frac{\sum_{s=1}^{\#\mathcal{C}_i}\sum_{t=1}^{\#\mathcal{C}_j} n_{s, t}\log \left(
\frac{n\cdot n_{s, t}}{n^i_s\cdot n^j_t}\right)}{\sqrt{\left(\sum_{s=1}^{\#\mathcal{C}_i}
n_s\log \left( \frac{n_s^i}{n}\right)\right)\left(\sum_{t=1}^{\#\mathcal{C}_j} n_t\log \left(
\frac{n_t^j}{n}\right)\right)}},
\]
where $\#\C_i$, $\#\C_j$ denote the number of clusters in the clusterings $\C_i$ and $\C_j$,
respectively. We note that $\phi^{\text{NMI}}(\mathcal{C}_i, \mathcal{C}_j)$ is invariant
under relabeling, as required above. If either $\C_i$ or $\C_j$ only contains a single
cluster, then the corresponding entropy is zero, and we define $\phi^{\text{NMI}}
(\mathcal{C}_i, \mathcal{C}_j)$ to be zero as well.

Given an ensemble of $m$ clusterings $\Lambda = \{\C_1,\ldots, \C_m\}$, in order to compare
the information that a single clustering $\C$ (not necessarily part of the ensemble) shares
with the ensemble $\Lambda$, the averaged normalized mutual information (ANMI) is defined
\cite[see][]{strehl2002cluster} as

\[
\phi^{\text{ANMI}}(\mathcal{C}, \Lambda) = \frac{1}{m}\sum_{i=1}^m \phi^{\text{NMI}}
(\mathcal{C}, \mathcal{C}_i).
\]
For easier notation, we refer to $\phi^{\text{NMI}}$ and $\phi^{\text{ANMI}}$ simply by
NMI and ANMI respectively.

\subsection{Consensus Clustering}\label{sec:cons}
The idea of consensus clustering (also termed ensemble clustering) is to generate a set of $m$
initial clusterings $\Lambda = \{\mathcal{C}_1, \dots, \mathcal{C}_m\}$ of some data set
and obtain a final clustering result $\mathcal{C}^*$ by integrating the initial results
\cite[see][]{vega2011survey,ghosh2011cluster}. As the name suggests, $\mathcal{C}^*$ should
represent a ``consensus'' or common denominator among all given initial clusterings from the
ensemble. Therefore one would want to
aim for a consensus clustering $\mathcal{C}^*$ with a high ANMI compared to the ensemble of
clusterings, that is
\begin{equation}
\label{eq:ANMI}
\mathcal{C}^* = \argmax_\C \phi^{\text{ANMI}}(\mathcal{C}, \Lambda),
\end{equation}
where the maximum is taken over a class of conceivable clusterings of the data.

The introduction of the consensus clustering strategy is motivated by two issues that are
notorious for existing clustering techniques: (i) Different clustering algorithms make
different assumptions about the structure of the data, rendering the right choice of algorithm
difficult if that structure is not known, (ii) it is difficult to choose hyperparameters for
those algorithms that have them.

Generating a set of initial clusterings $\Lambda$ can be done either by using one classical
clustering
algorithm with various hyperparameter settings or different initializations, or by clustering
the data using
different
clustering algorithms. It is advisable to use clustering
algorithms that work well for the intrinsic structure of the data if it is known beforehand.
Otherwise, a variety of clustering algorithms may be used so that the consensus clustering
integrates as much information as possible.

\subsection{Computation of consensus clusterings}
\label{sec:consensuscompu}
Computing the consensus clustering by direct brute force maximization of \eqref{eq:ANMI}  is
not tractable since the number of possible consensus labellings grows exponentially in the
number of data points. On the other hand, greedy algorithms are computationally not feasible
and lead to local maxima according to \citet{strehl2002cluster}. Therefore,
one needs to find good heuristics and rather use \eqref{eq:ANMI} as an evaluation measure.
Several approaches for finding the consensus clustering $\mathcal{C}^*$ are discussed in the
literature \citep{vega2011survey,xu2012analysis}. In this paper, we give two
possibilities \citep{strehl2002cluster}.

\paragraph{Reclustering points.} This approach uses some distance
\[
d_p:\{x_1, \dots, x_n\} \times \{x_1, \dots, x_n\}\to \mathbb{R}^+_0
\]
on the data points in the following way: For each pair of points $(x_i, x_j)$ one can count
the number of clusterings from $\Lambda$, in which $x_i$ and $x_j$ have a different label.
That is, $d_p$ is the Hamming distance \citep{hamming1950error} on the points in this case.
Then to get $\mathcal{C}^*$, one reclusters the points with a similarity based clustering algorithm,
i.e. one that only requires a distance metric between the points as input, opposed to
coordinates in Euclidean space. Examples include the agglomerative hierarchical clustering
method \citep{ward1963hierarchical} and spectral clustering \citep{ng2002spectral}.

\paragraph{Meta-clustering.} Another possibility is to introduce a distance $d_C$ on the set
$\{C \in \C_l: l=1, \dots m\}$
containing all clusters from the clusterings of the ensemble  and to cluster these into
``meta-clusters''.
Therefore, any data point $x$ is appearing $m$ times in this ``meta-clustering'' and is
finally assigned to
the meta-cluster, that it belongs most often to.\\
\\
Although the second approach should be
computationally more efficient than the first approach, one needs to
implement a graph partitioning algorithm, if one follows \citet{strehl2002cluster}.
The first approach on the
other hand is easier to understand and implement, and thus we choose this method for our
experiments in
Section \ref{sec:results} and employ hierarchical clustering as the similarity based
clustering method.
Note that the focus of this paper is not the choice of the consensus function, but to present
a framework for
finding hyperparameters of clustering algorithms.

There is one more subtlety to computing the
consensus clustering using this approach: Most consensus
clustering methods require the practitioner to know the number of clusters $k^*$ for the
computation of $\mathcal{C}^*$ in advance. Anyhow, our aim is to find clustering algorithm
hyperparameters and not exchanging one choice for another. Therefore, we will show in an
example, that for large enough $k^*$, the hyperparameters determined by our algorithm using
consensus clustering is invariant under the choice of $k^*$, so that this does not pose a
problem. Moreover, there are heuristics to choose $k^*$ in a principled way. See Sections
\ref{sec:hyperparam_search} and \ref{sec:synth_data_fuzzy} for more details.

\section{Hyperparameter Search using NMI and ANMI}\label{sec:hyperparam_search}

Our proposed approach for finding a reasonable choice of hyperparameters for some given
clustering algorithms is twofold:
First, an ensemble of clusterings $\Lambda$ is produced by varying the hyperparameters and
clustering algorithms considered. Second, the point in parameter space is identified which
produced the clustering that shared the most information given by the
ensemble $\Lambda$. We will consider two strategies for finding this clustering.
For a visual representation of the two approaches, see Figure \ref{fig: schemata}.

For the sake of easier notation, we view the choice of a clustering algorithm as a
hyperparameter as well.
Having that in mind, let $S$ denote the finite set of all
considered hyperparameter configurations, $\mathcal{C}(s)$ the clustering according to the
hyperparameter configuration $s\in S$ and let $\Lambda(S)=\{\mathcal{C}(s): s\in S\}$ be the
set of all clusterings depending on $S$.
The simplest choice for $S$ is a grid search such that $S=P^{(1)}\times \dots \times P^{(l)}$
where $P^{(i)}$ denotes the range of hyperparameter $i$.

\paragraph{\emph{Strategy 1}: ANMI maximization.} Optimal hyperparameters are selected according to
their ANMI (averaged normalized mutual information) score. The ``ANMI-best'' hyperparameter
configuration $s^{\text{ANMI}}$ is thus defined as
\[
s^{\text{ANMI}} = \argmax_{s\in S} \phi^{\text{ANMI}}(C(s),\Lambda(S)\setminus \{C(s)\}).
\]

\paragraph{\emph{Strategy 2}: Best match with consensus clustering.} Rather than aggregating
information over $\Lambda(S)$ with the ANMI score, we aggregate information over $\Lambda(S)$
by constructing a consensus clustering $\mathcal{C}^*$ from $\Lambda(S)$ (as described in
Section \ref{sec:consensuscompu}). We then select optimal hyperparameters according to their
NMI score with $\mathcal{C}^*$, that is
\[
s^{*} = \argmax_{s\in S} \phi^{\text{NMI}}(\mathcal{C}(s), \mathcal{C}^*).
\]
This idea translates into the following algorithm \ref{alg:param_search}.

\begin{algorithm}[!htb]
\SetAlgoLined
\DontPrintSemicolon
\KwIn{Set $S$ of hyperparameter configurations for chosen clustering algorithms; consensus clustering algorithm with chosen hyperparameters; data $D=\{x_1, \dots, x_n\}$ to cluster}
\KwOut{Optimal (in the sense of the NMI-criterion) hyperparameter configuration $s^*\in S$}
\ForEach{$s\in S$}{
Produce clustering $\mathcal{C}(s)$ with hyperparameter configuration $s$ and with the corresponding clustering algorithm}
Produce consensus clustering $\mathcal{C}^*$ based on $\Lambda(S)$\;
\Return $s^*=\argmax_{s\in S} \phi^{\text{NMI}}(\mathcal{C}(s), \mathcal{C}^*)$\;
\caption{Hyperparameter search using \emph{Strategy 1}}
\label{alg:param_search}
\end{algorithm}

Note that one might also use other normalizations for measuring mutual information or
different external clustering similarity measures like the adjusted Rand index $ARI(\C,\C')$.
See Subsection \ref{sec:real_data} for an experimental comparison.

When applying \emph{Strategy 2} we need to compute the consensus clustering. Often the
hyperparameter $k^*$ is needed as an input, i.e. the number of clusters that have to be chosen
for computing the consensus clustering. Since we want to find hyperparameters for a clustering
method, we have to address the problem of this additional hyperparameter. In
\ref{sec:synth_data_fuzzy} our clustering experiments indicate that \emph{Strategy 2} is very robust
regarding the choice of $k^*$, when $k^*$ is at least large enough. Therefore, just choosing a
large enough $k^*$ should suffice. However, there are even more ways to avoid this
restriction, one possible solution is to
use the PAC (proportion of ambiguous clustering) measure
\citep{senbabaouglu2014critical,senbabaoglu2014reassessment,monti2003consensus} to infer the
optimal $k^*$ by minimizing PAC over clusterings from the ensemble with different $k$. This
can be performed as a preparation step for \emph{Strategy 2}.

\paragraph{Computational complexity.} Considering their computational complexity, the two
strategies differ. Both algorithms might not be feasible in all situations. Denoting the
number of data points with $n$, the typical number of clusters with
$k$
and the number of clusterings in the ensemble with $m$, we can do a na\"ive computational
complexity analysis. For computing our consensus function, we computed the Hamming distance
for every pair of data points, which is $O(n^2m)$. Using agglomerative hierarchical clustering
is $O(n^2\log(n))$ in general. Moreover, for one NMI-evaluation, we need to go through all
data points to count cluster label pairs, which is $O(n)$, and then add $O(k^2)$ terms in the
enumerator, which, for $m$ clusterings in the ensemble, yields $O(mn + mk^2)$. In total,
algorithm \ref{alg:param_search} costs $O(n^2\log(n) + mn +k^2m)$. In contrast, finding the
ANMI maximizing (\emph{Strategy 1}) clustering costs one NMI-evaluation for every pair of the $m$ clusterings,
i.e. $O(nm^2)$. In summary, algorithm \ref{alg:param_search} scales worse in the number of
data points, whereas finding the ANMI maximizing (\emph{Strategy 1}) clustering scales worse in the number of
clusterings in the ensemble.

\section{Experimental Results}\label{sec:results}

We will give some examples of applying the strategies
proposed in the previous section on two synthetic data sets and one real-world data set.
Namely we apply the algorithm for finding the best hyperparameter configuration on a very
fuzzy data set, see Section \ref{sec:synth_data_fuzzy} and a data set of handwritten digits in
Section \ref{sec:real_data}. Moreover we present the results of using the algorithm for
finding the best clustering method on a synthetic data set in Section
\ref{sec:synth_data_spiral}. In each case %the ``NMI-best'' and ``ANMI-best'' clustering
the resulting clustering of the two strategies can be compared to the consensus clustering.
For the algorithmic experiments, we implemented the methods in Python relying heavily on the
Scikit-Learn library \citep{scikit-learn}.

For testing our proposed methods, we perform the following steps: First, we choose
one or several classical clustering algorithms and hyperparameter configurations for each,
i.e. we set $S$ in order to generate our set of initial clusterings $\Lambda(S)$ on a
grid. Second, we construct the consensus clustering as described in Section
\ref{sec:consensuscompu} using the Hamming distance and hierarchical clustering. Next, the NMI
of each clustering from $\Lambda(S)$ with the consensus clustering is evaluated. We
additionally calculate the ANMI of the consensus clustering as well as of each clustering with
the remaining clusterings. The average normalized mutual information of a clustering with a
set of clusterings is one heuristic to measure how good they agree, i.e. it measures the
shared information between the clustering and set of clusterings. Thus it would be desirable
for the consensus clustering to have a high ANMI with $\Lambda(S)$. And last, we find the
%``NMI-best''
hyperparameter configuration
of the ANMI maximization (\emph{Strategy 1}) and of the best match with the consensus clustering
(\emph{Strategy 2})
and analyze the results.

\subsection{Experiments on synthetic data}\label{sec:synth_data}

We chose two very different artificial data sets. For the ``fuzzy'' data set we estimate the best DBSCAN hyperparameters in the
following. However, for the ``spiral'' data set we
will estimate not only the best hyperparameters but also test which clustering algorithm
should be preferred.

\subsubsection{Finding hyperparameters for DBSCAN: The fuzzy data
set}\label{sec:synth_data_fuzzy}

%%%%%%%%%%%%%%%%%%%%%%%%%%%%%%%%%%%%%%%%%%%%%%%%%%%%%%%%
%FUZZY - DBSCAN
%%%%%%%%%%%%%%%%%%%%%%%%%%%%%%%%%%%%%%%%%%%%%%%%%%%%%%%%

\begin{figure}[!htb]
\centering
        \begin{subfigure}[b]{0.32\textwidth}
            \centering
            \includegraphics[width=\textwidth]{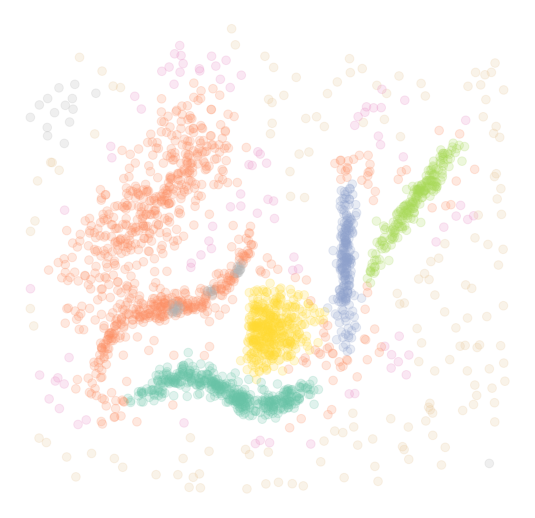}
            \caption{Consensus clustering $\mathcal{C}^*$,\\ ANMI = 0.356}\label{fig:fuzzy_dbscan_consensus}
        \end{subfigure}
        \begin{subfigure}[b]{0.32\textwidth}
            \centering
            \includegraphics[width=\textwidth]{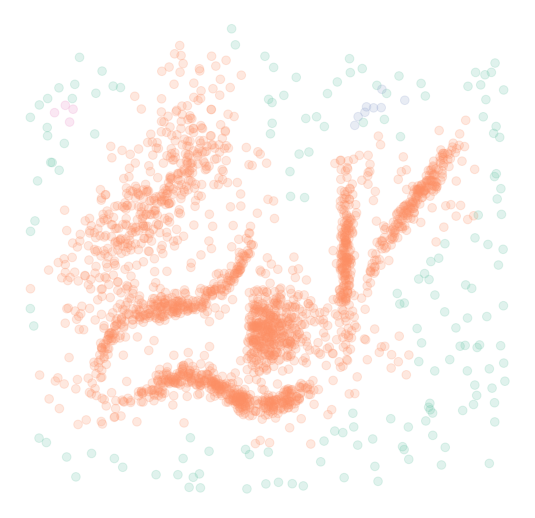}
  \caption{\emph{Strategy 1} clustering,\\ ANMI = 0.372 , NMI = 0.38}\label{fig:fuzzy_dbscan_ANMI}
        \end{subfigure}
                \begin{subfigure}[b]{0.32\textwidth}
            \centering
            \includegraphics[width=\textwidth]{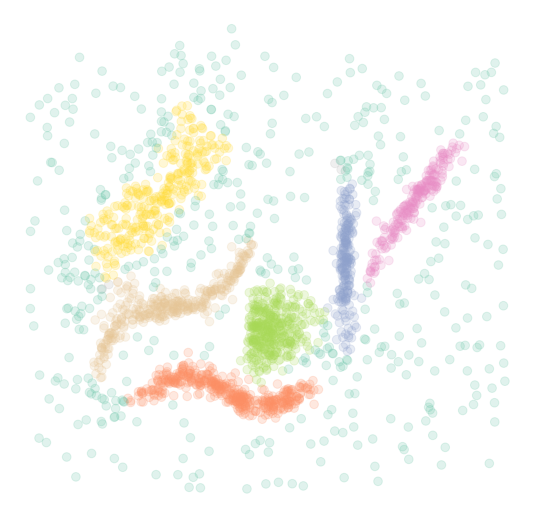}
  \caption{\emph{Strategy 2} clustering,\\ ANMI = 0.292, NMI = 0.813 }\label{fig:fuzzy_dbscan_NMI}
        \end{subfigure}
        \caption{The ``fuzzy'' synthetic data set contains fuzzy, non-blob-shaped clusters and
many noisy data points. The clusterings chosen by \emph{Strategy 1} and \emph{Strategy 2} over a grid of
clusterings (see Figure \ref{fig:fuzzy_dbscan_all}) are shown here. }\label{fig:fuzzy_results}
\end{figure}

\begin{figure}[!htb]
\centering
\includegraphics[width=\textwidth]{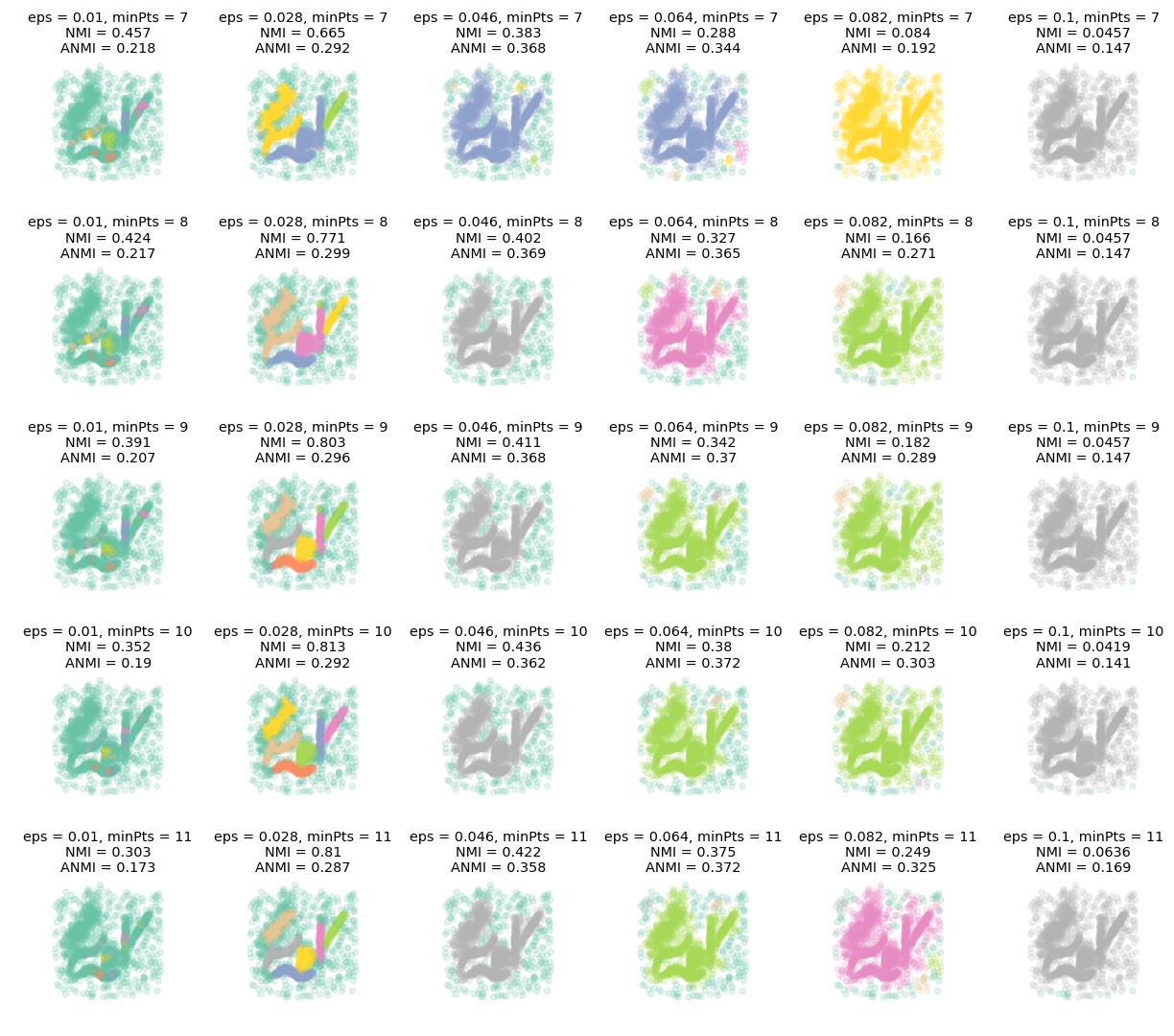}
\caption{DBSCAN clusterings over a grid of hyperparameters for the ``fuzzy'' data set. We
varied $\varepsilon$ and $m_{\mathrm{points}}$ to produce $\Lambda(S)$ on a grid. In order to
generate the consensus clustering, the number of clusters has to be set, here we set $k^* =
10$. For each clustering we computed the NMI with the consensus clustering (see Figure
\ref{fig:fuzzy_results}) and the ANMI with the set of clusterings.}
\label{fig:fuzzy_dbscan_all}
\end{figure}

The ``fuzzy'' data set \citep{fuzzydata} contains some non-globular clusters and some noise and
is therefore not easy to cluster for most clustering algorithms. The data set consists of 2309
two-dimensional data points.
To generate the set of initial clusterings $\Lambda$ we used the DBSCAN clustering algorithm
with a range of different hyperparameters (in particular we varied $\varepsilon$ and
$m_{\mathrm{points}}$).
Density Based Spatial Clustering of Applications with Noise (DBSCAN) was first proposed by
\citet{ester1996density} and is the most commonly used density based clustering approach in
metric spaces. Given a density specification consisting of a radius hyperparameter
$\varepsilon$ and a minimal point count hyperparameter $m_{\mathrm{points}}$, the algorithm
iterates over the data set and searches clusters of density-connected  structures, i.e. areas
that satisfy a number of at least $m_{\mathrm{points}}$ in a $\varepsilon$-neighborhood of
given cluster members and expanding the detected clusters successively.
DBSCAN is very sensitive to its hyperparameters, but if they are well chosen, it is capable
of detecting highly non-convex, densely connected structures in the data.

Having constructed the grid of clusterings (see Figure \ref{fig:fuzzy_dbscan_all}), it is not
surprising that many clusterings are not very reasonable as DBSCAN is very sensitive to its
hyperparameters. The consensus clustering should be a common denominator of all the
clusterings. Given $\Lambda(S)$, the consensus clustering $\mathcal{C}^*$ can be produced and
we can find the best match (\emph{Strategy 2}) hyperparameter configuration and resulting clustering (see Figure \ref{fig:fuzzy_results}).

Surprisingly the consensus clustering doesn't look as good as the clustering with the best
match with $\C^*$ (\emph{Strategy 2}): For example in the consensus clustering, two obviously
separate clusters are not clearly distinguished.
A possible explanation for this effect is the influence of the substantial number of unreasonable clusterings in the ensemble on $\mathcal{C}^*$ (see Figure \ref{fig:fuzzy_dbscan_all}). Nevertheless, already minor indications for distinguishing
these two clusters in $\mathcal{C}^*$ are enough to increase the NMI-score of the ``correct''
clustering. Note that NMI also accounts for the size of the clusters.

The results of our algorithms depend on the initially constructed ensemble of clusterings. In
the given case, many clusterings in the ensemble contain only one or two clusters, i.e.
either the whole data set is one cluster or the structure is one cluster, the noise the other
cluster. Therefore the ANMI maximizing clustering (\emph{Strategy 1}) resembles these two
observations (see Figure \ref{fig:fuzzy_results}).

\begin{figure}[!htb]
\centering
\includegraphics[width=\textwidth]{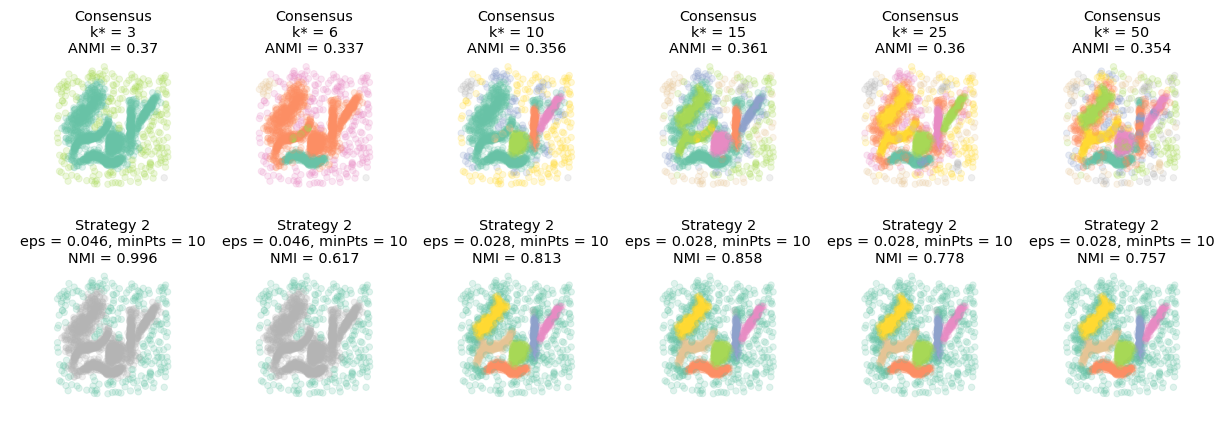}
\caption{Consensus clusterings and the best match (\emph{Strategy 2}) clusterings for different values of the
number of clusters in the consensus clustering $k^*\in \{3, 6, 10, 15, 25, 50\}$. For each
\emph{Strategy 2} clustering, the found hyperparameters $s^*$ are given.
The clustering
ensemble is the same DBSCAN ensemble as in Figure \ref{fig:fuzzy_dbscan_all}. If $k^*$ is
chosen to be large enough, the choice of the clustering always remains the same.}
\label{fig:fuzzy_dbscan_k-robustness}
\end{figure}

The experiments in Figure \ref{fig:fuzzy_results} and \ref{fig:fuzzy_dbscan_all} used the a priori choice of $k^* = 10$ for the number of clusters in $\mathcal{C}^*$. In Figure \ref{fig:fuzzy_dbscan_k-robustness} we repeat the same experiments but vary $k^*\in \{3, 6, 10, 15, 25, 50\}$. We note that both the reported best match (\emph{Strategy 2}) clustering as well as the chosen hyperparameters found by \emph{Strategy 2} are consistent for all $k^* \geq 10$. Looking at the NMI-values for the most reasonable four clusterings in the ensemble, even the
second best clustering remains the same for large enough $k^*$.

\subsubsection{Finding the appropriate algorithm plus hyperparameters: The spiral data set}
\label{sec:synth_data_spiral}

%%%%%%%%%%%%%%%%%%%%%%%%%%%%%%%%%%%%%%%%%%%%%%%%%%%%%%%%
%SPIRAL ALGS
%%%%%%%%%%%%%%%%%%%%%%%%%%%%%%%%%%%%%%%%%%%%%%%%%%%%%%%%

\begin{figure}[!htb]
\centering
\small
        \begin{subfigure}[b]{0.32\textwidth}
            \centering
            \includegraphics[width=\textwidth]{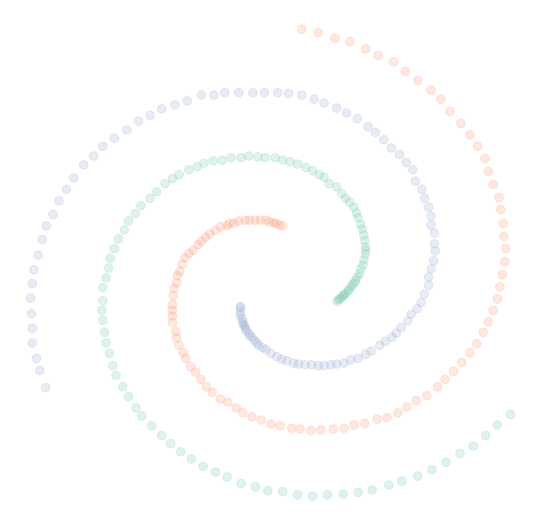}
            \caption{Consensus clustering $\mathcal{C}^*$,\\ ANMI = 0.58 }\label{fig:spiral_algs_consensus}
        \end{subfigure}
        \begin{subfigure}[b]{0.32\textwidth}
            \centering
            \includegraphics[width=\textwidth]{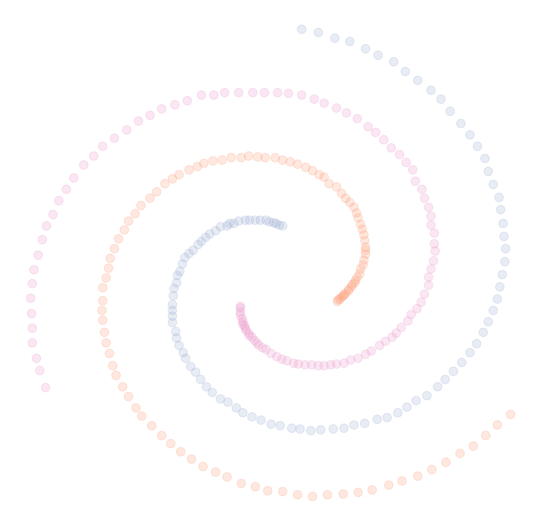}
  \caption{\emph{Strategy 1} clustering,\\ ANMI = 0.58 , NMI = 1.0}\label{fig:spiral_algs_ANMI}
        \end{subfigure}
        \begin{subfigure}[b]{0.32\textwidth}
            \centering
            \includegraphics[width=\textwidth]{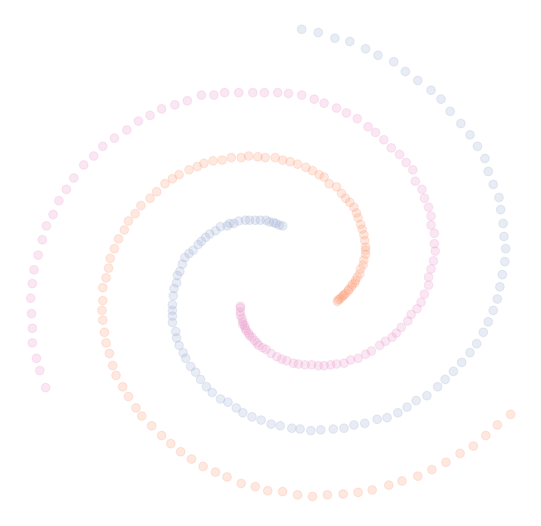}
  \caption{\emph{Strategy 2} clustering,\\ ANMI = 0.58 , NMI = 1.0 }\label{fig:spiral_algs_NMI}
        \end{subfigure}
        \caption{Synthetic spiral data set embedded in 2D. The data points form three spiral
arms corresponding to three clusters. In this example, the results of our algorithms all lead
to the same clustering as the consensus clustering, see Figure \ref{fig:spiral_algs_all} for
the grid of clusterings. }
        \label{fig:spiral_results}
\end{figure}

\begin{figure}[!htb]
\centering
\includegraphics[width=\textwidth]{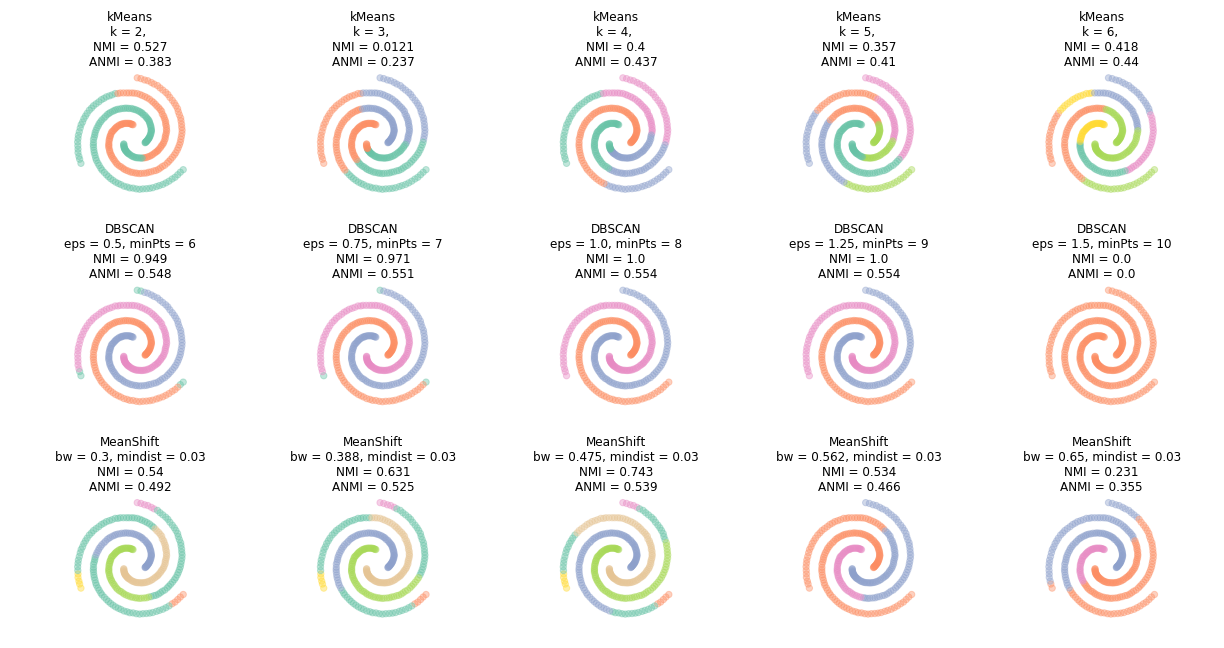}
\caption{Clusterings over a grid of different algorithms for the spiral data set, namely for
(i) k-Means (first row), (ii) DBSCAN (second row) and, (iii) Mean Shift (third row).
(i) It is expected that k-Means performs badly, the NMI with $\mathcal{C}^*$ is
quite low.
On the other hand (ii) DBSCAN does very well even for dissimilar hyperparameters. Two
clusterings  on the grid even share all information with $\mathcal{C}^*$.
And in the last row (iii), the Mean Shift algorithm outperforms k-Means, but doesn't produce
an outcome as good as DBSCAN.
For producing $\mathcal{C}^*$ the number of clusters was set to $k^*=3$.}
\label{fig:spiral_algs_all}
\end{figure}

The spiral data set \citep{ClusteringDatasets} contains three spiral arms and no noise or
outliers. It consists of 312 data points in $\mathbb{R}^3$. The difficulty lies in the fact that the spiral arms, which form
the clusters, are elongated and not blob-shaped at all.
We chose to use k-Means \citep[see][]{lloyd1982least,macqueen1967some}, DBSCAN and the Mean
Shift algorithm over a range a hyperparameters to generate the set of clusterings
$\Lambda(S)$ (see Figure \ref{fig:spiral_algs_all}). The k-Means algorithm works well on
blob-shaped data, thus one wouldn't expect that it produces reasonable clusterings for any
hyperparameter configuration here. Both DBSCAN and Mean Shift are density-based and thus
could give good results on the spiral data set.
The Mean Shift clustering approach builds on the idea that the data points represent samples
from some underlying probability density function. It is an iterative algorithm in which the
data points are shifted in the direction of maximum increase in the data sets' density until
convergence \citep{comaniciu2002mean}. Choosing a sensible bandwidth hyperparameter is
essential for a good clustering outcome.

To generate $\Lambda(S)$ we varied  the number of clusters $k$ for K-Means, for DBSCAN we
used different values of $\varepsilon$ and $m_{\mathrm{points}}$ and for the Mean Shift
algorithm we varied the bandwidth and minimum distance. The produced grid of clusterings and
hyperparameter configurations can be seen in Figure \ref{fig:spiral_algs_all}. As expected
k-means does not produce good clustering results on this type of data, neither does the Mean
Shift algorithm for the chosen hyperparameter configurations, but DBSCAN produces very
satisfying results.

The consensus clustering combines the clusterings from $\Lambda(S)$ and is able to
distinguish the three spiral arms as three clusters, see Figure
\ref{fig:spiral_results}. The two satisfying clustering results generated using DBSCAN have
NMI 1.0 with the consensus, i.e. they are exactly the same. Thus the best match (\emph{Strategy 2})
hyperparameter configurations will give the same result as the consensus clustering. In this
example, also the ANMI maximizing (\emph{Strategy 1}) clustering is the same clustering as the consensus
clustering. Our two algorithms thus yield the same clustering result.

\subsection{Experiments on real-world data: The digit data set} \label{sec:real_data}

\label{sec:digits}

\begin{figure}[!htb]
    \centering
	\includegraphics[width=0.5\textwidth]{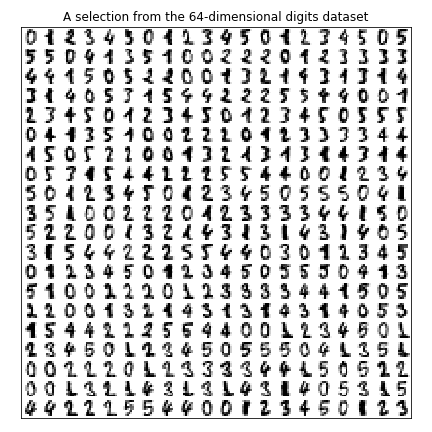}
	\caption{The digits data set \citep{Lichman:2013} consists of preprocessed $8\times 8$ images, thus each data point is 64-dimensional. For visual verification purposes
with 2D embeddings, we only used the handwritten digits $0$ to $5$ of which there are $1038$.}
   \label{fig:digits}
\end{figure}

%%%%%%%%%%%%%%%%%%%%%%%%%%%%%%%%%%%%%%%%%%%%%%%%%%%%%%%%
%DIGITS - DBSCAN
%%%%%%%%%%%%%%%%%%%%%%%%%%%%%%%%%%%%%%%%%%%%%%%%%%%%%%%%

\begin{figure}[!htb]
\centering
        \begin{subfigure}[b]{0.32\textwidth}
            \centering
            \includegraphics[width=\textwidth]{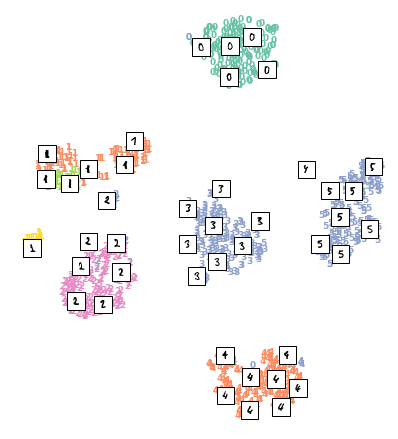}
            \caption{Consensus clustering $\mathcal{C}^*$,\\ ANMI = 0.188}
            \label{fig:mnist_dbscan_consensus}
        \end{subfigure}
        \begin{subfigure}[b]{0.332\textwidth}
            \centering
            \includegraphics[width=\textwidth]{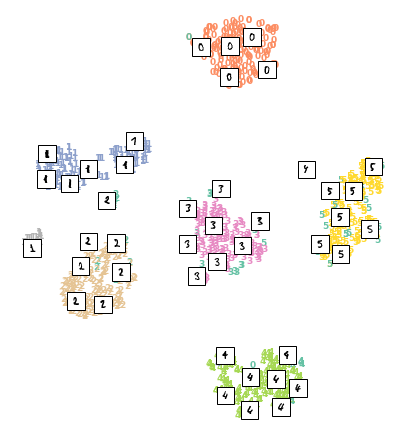}
  \caption{\emph{Strategy 1} clustering,\\ ANMI = 0.193 , NMI = 0.816}
  \label{fig:mnist_dbscan_ANMI}
        \end{subfigure}
        \begin{subfigure}[b]{0.32\textwidth}
            \centering
            \includegraphics[width=\textwidth]{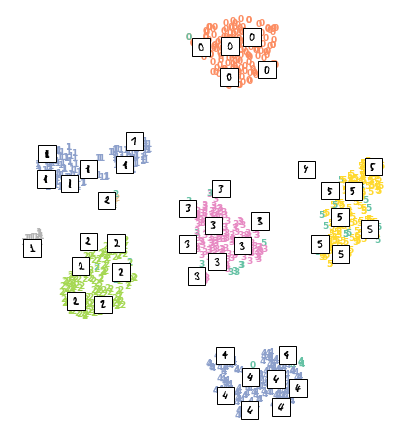}
  \caption{\emph{Strategy 2} clustering,\\ANMI = 0.188, NMI = 0.834}
  \label{fig:mnist_dbscan_NMI}
        \end{subfigure}
        \caption{Results of our methods on a grid of different hyperparameters for DBSCAN. The digits data set
        is embedded in 2D using the t-distributed Stochastic Neighbor Embedding. For the underlying clustering ensemble, we refer the reader to Figure \ref{fig:digits_dbscan_all}.}
        \label{fig:mnist_dbscan_results}
\end{figure}

\begin{figure}[!htb]
\centering
\includegraphics[width=\textwidth]{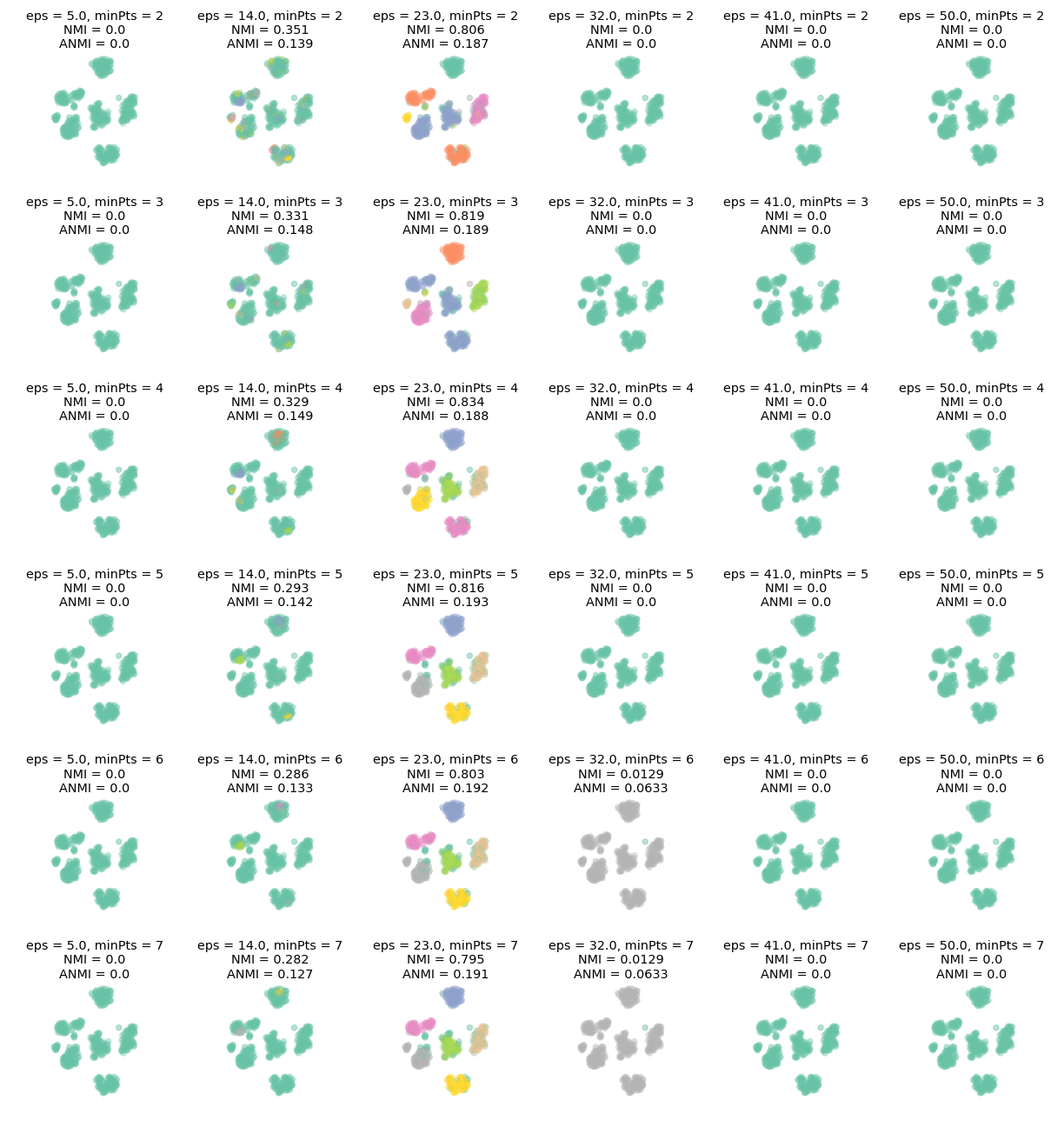}
\caption{Grid of clusterings for the digits data over a range of DBSCAN hyperparameters. To produce the consensus clustering, the number of clusters was set to $k^*=6$. The digits data set is embedded in 2D using the t-distributed Stochastic Neighbor Embedding.}
\label{fig:digits_dbscan_all}
\end{figure}

%%%%%%%%%%%%%%%%%%%%%%%%%%%%%%%%%%%%%%%%%%%%%%%%%%%%%%%%
%DIGITS - TABLE
%%%%%%%%%%%%%%%%%%%%%%%%%%%%%%%%%%%%%%%%%%%%%%%%%%%%%%%%

\begin{table}
\vspace{-1cm}
\thisfloatpagestyle{empty}
\centering
\small
\scalebox{0.9}{
\begin{tabular}{l | rrrrr |r}

\toprule
       &  $\varepsilon=5.0$  &  $\varepsilon=14.0$ &   $\varepsilon=23.0$ &  $\varepsilon=32.0$ &  $\varepsilon=41.0$ & Hyperpar.\\
\midrule
ANMI &   0.0 &  0.14 &   0.19 &  0.00 &   0.0 &  \\
ARI &   0.0 &  0.12 &   0.67 &  0.00 &   0.0 &   \\
CHI &   --$^*$ &  4.34 &  44.99 &   -- &   -- &  \\
DI1 (max.) &   -- &  0.18 &   \textcolor{red}{0.32} &   -- &   -- &  $m_{\mathrm{points}}=2$\\
DI2 &   -- &  0.34 &   0.55 &   -- &   -- &  \\
NMI &   0.0 &  0.35 &   0.81 &  0.00 &   0.0 &  \\
Silhouette &   -- & -0.30 &   0.04 &   -- &   -- &  \\ \hline
ANMI &   0.0 &  0.15 &   0.19 &  0.00 &   0.0 &  \\
ARI &   0.0 &  0.10 &   0.68 &  0.00 &   0.0 &  \\
CHI &   -- &  6.75 &  66.25 &   -- &   -- &  \\
DI1 &   -- &  0.14 &   0.19 &   -- &   -- &  $m_{\mathrm{points}}=3$\\
DI2 &   -- &  0.41 &   0.78 &   -- &   -- &  \\
NMI &   0.0 &  0.33 &   0.82 &  0.00 &   0.0 &  \\
Silhouette &   -- & -0.33 &   0.11 &   -- &   -- & \\ \hline
ANMI &   0.0 &  0.15 &   0.19 &  0.00 &   0.0 &  \\
ARI (max.) &   0.0 &  0.09 &   \textcolor{red}{0.69} &  0.00 &   0.0 &  \\
CHI &   -- & 10.46 & 101.43 &   -- &   -- &  \\
DI1 &   -- &  0.14 &   0.23 &   -- &   -- &  $m_{\mathrm{points}}=4$\\
DI2 &   -- &  0.39 &   0.93 &   -- &   -- &  \\
NMI (max.) &   0.0 &  0.33 &   \textcolor{red}{0.83} &  0.00 &   0.0 &  \\
Silhouette &   -- & -0.26 &   0.17 &   -- &   -- & \\ \hline
ANMI (max.) &   0.0 &  0.14 &   \textcolor{red}{0.19} &  0.00 &   0.0 &  \\
ARI &   0.0 &  0.07 &   0.60 &  0.00 &   0.0 &  \\
CHI (max.) &   -- & 10.69 & \textcolor{red}{123.87} &   -- &   -- &  \\
DI1 &   -- &  0.12 &   0.20 &   -- &   -- &  $m_{\mathrm{points}}=5$\\
DI2 &   -- &  0.42 &   0.92 &   -- &   -- &  \\
NMI &   0.0 &  0.29 &   0.82 &  0.00 &   0.0 &  \\
Silhouette &   -- & -0.26 &   0.20 &   -- &   -- &  \\ \hline
ANMI &   0.0 &  0.13 &   0.19 &  0.06 &   0.0 &  \\
ARI &   0.0 &  0.06 &   0.59 & 0.00 &   0.0 &  \\
CHI &   -- & 12.65 & 123.35 &  2.34 &   -- &  \\
DI1 &   -- &  0.12 &   0.20 &  0.21 &   -- &  $m_{\mathrm{points}}=6$\\
DI2 &   -- &  0.40 &   0.92 &  1.08 &   -- &  \\
NMI &   0.0 &  0.29 &   0.80 &  0.01 &   0.0 & \\
Silhouette &   -- & -0.18 &   0.20 &  0.06 &   -- &  \\ \hline
ANMI &   0.0 &  0.13 &   0.19 &  0.06 &   0.0 &  \\
ARI &   0.0 &  0.05 &   0.58 & 0.00 &   0.0 &  \\
CHI &   -- & 21.55 & 122.83 &  2.34 &   -- &  \\
DI1 &   -- &  0.12 &   0.20 &  0.21 &   -- &
$m_{\mathrm{points}}=7$\\
DI2 (max.) &   -- &  \textcolor{red}{0.94} &   0.93 &  1.08 &   -- &  \\
NMI &   0.0 &  0.28 &   0.79 &  0.01 &   0.0 & \\
Silhouette (max.) &   -- & -0.12 &   \textcolor{red}{0.20} &  0.06 &   -- & \\
\bottomrule
\end{tabular}
}
\caption{A selection of commonly used cluster validation metrics on the
clusterings depicted in
Figure \ref{fig:digits_dbscan_all} (rounded values). The maximum value for each metric over the range of hyperparameters is highlighted in red. Definitions of the used measures can be found in Appendix A.\\
$^*$value undefined due to single detected cluster
}
\label{tab:digits}
\end{table}

The digit data set \citep{Lichman:2013} contains labeled handwritten digits and consists of
preprocessed $8\times 8$ pixel images (see Figure \ref{fig:digits}
for some samples). For visual verification purposes, we used the t-distributed Stochastic
Neighbor Embedding \citep{maaten2008visualizing} to display a clustered subset of the data
(labels corresponding to the digits zero to five)
in two dimensions.
We want to emphasize that the actual clustering was done in the original space of $8\times 8$
pixels, i.e. \emph{not} in the space obtained by t-SNE.

We used the DBSCAN algorithm (see Figure \ref{fig:digits_dbscan_all})
with a grid of different $\varepsilon$- and $m_\text{points}$-parameters, produced the
consensus clustering and calculated the NMI and ANMI for every clustering.

The generated consensus as well as the outcomes of \emph{Strategy 1} (ANMI maximization)
and \emph{Strategy 2} (best match clustering) are shown in Figure \ref{fig:mnist_dbscan_results}.
The consensus clustering fails to separate the digits ``3'' and ``5'' in the data (dark blue).
Interestingly,
the \emph{Strategy 2} clustering distinguishes in between
the ``3'' and ``5'' (dark blue), but fails to
separate the ``1'' and ``4''. The ANMI-maximized (\emph{Strategy 1}) clustering is able to capture the main
classes in the data up to small noise, even though the consensus is not classifying
everything correctly.

Moreover, we compare the results to a choice of classical
internal clustering validation measures, including Dunn index-type measures
\citep[DI, see][]{dunn74},
the silhouette coefficient \citep{rousseeuw87silhouette} and
the Cali\'nski-Harabasz index \citep[CHI, ][]{calinski74}, as well as
external measures like the Rand index
\citep[ARI, see][]{rand71},
see Appendix A for detailed definitions.
The Rand index is taken with respect to the computed consensus clustering.
From the detailed listing of the measure values in Table \ref{tab:digits} we conclude
that in this example maximizing other classical clustering measures on
the ensemble would not always have the same effect as maximizing NMI (\emph{Strategy 2}) or
ANMI (\emph{Strategy 1}). Both \emph{Strategy 1} and 2 result in similar hyperparameter choices that are also chosen by
the  Cali\'nski-Harabasz index, respectively by the adjusted Rand index.
In some situations (e.g. DI1, DI2),
it might even be misleading for the user when picking the inappropriate clustering measure.

It is interesting to note that the same intrinsic measures on $\mathcal{C}^*$ (see Table  \ref{tab:digits_results}) can either be better than all the values
observed in
the ensemble (DI2),
mediocre (CHI) or significantly worse (Silhouette). Thus, $\mathcal{C}^*$ itself can often be a poor clustering decision. However, \emph{Strategies 1} and \emph{2} do lead to competitive clusterings in terms of most of the measures considered here, with \emph{Strategy 1} slightly outperforming \emph{Strategy 2}.

\begin{table}[tbp]
\centering

\begin{tabular}{l | ccc}
\toprule
 & Consensus $\mathcal{C}^*$  & \emph{Strategy 1} & \emph{Strategy 2}  \\ \midrule
 CHI & 113.42  & 123.87 & 101.43\\
 DI & 0.20  & 0.20 & 0.23\\
 DI2 & 1.01  & 0.92 & 0.93\\
 Silhouette & 0.13  & 0.20 & 0.17\\ \bottomrule
\end{tabular}
\caption{Commonly used internal clustering metric values for
ANMI- and best match-maximizers (\emph{Strategy 1} and \emph{2} respectively) from Table \ref{tab:digits} compared to the consensus clustering $\mathcal{C}^*$.}
\label{tab:digits_results}
\end{table}

\section{Conclusion} \label{sec:conclusion}
In this paper, we introduced a technique to validate and compare elements in a clustering
ensemble
using aggregated common information about clusters in the ensemble.
This very general approach can be modified
using different consensus strategies and external clustering or classification accuracy
measures.
We may furthermore interpret this technique as a way of making use of external verification
methods even though a priori there is no external clustering we can compare the results to.
To do so, we produced external information by constructing an ensemble of clusterings with
different hyperparameters and clustering algorithms, such that the clustering ensemble is
diverse with respect to the clusterings.
And we make use of the information that is contained in the ensemble by evaluating
clusterings against it.

Using normalized mutual information and consensus clustering, as well as the average
normalized mutual information,
we applied the developed heuristic to artificial as well as real-world data sets in order to
\begin{itemize}
\item[(i)] implicitly
compare a clustering to all remaining clusterings in the ensemble,
\item[(ii)] search for optimal clustering hyperparameters for a given data set,
\item[(iii)] determine a reasonable clustering algorithm for a given problem.
\end{itemize}
We showed that this method is able to filter for clusterings containing the highest degree of information in a noisy ensemble. In our experiments, the clustering produced with the found optimal hyperparameters, was always better or as good as the consensus clustering $\mathcal{C}^*$. This might be due to the fact that ``correct'' structures are in general more likely to be persistent in multiple clusterings, than ``wrong'' structures. Therefore, the consensus clustering $\mathcal{C}^*$ picks up on the correct structures (though it is perturbed by bad clusterings to some extent), which causes good clusterings to have high NMI with $\mathcal{C}^*$.

In the DBSCAN examples, our techniques work particularly well, since for too small bandwidth parameters (large clusters), there is few information to share with the ensemble or the consensus. On the other hand, for too high bandwidth parameters, the information contained in one particular clustering is too detailed to appear in multiple clusterings.

% Possible applications (produce whole sentences):
% \begin{itemize}
% \item generally, one does know the clustering parameters, needs a strategy to find them
% \item in our experiments, clutering outcome with found parameters equal or better than
%consensus clustering
% \item knowing parameters helps to understand data better (e.g. knowing ``true'' $k$ is
%valuable or eps tells you, how dense the clusters are)
% \end{itemize}
Future research may be conducted in directions such as algorithm
complexity reduction (e.g. by using different consensus clustering methods) and
parallelization, hyperparameter search in terms of random sampling and application of other
quality measures
alongside NMI or ANMI.
Another question to be investigated is the robustness with respect to the number of clusters
given to the consensus clustering as a hyperparameter. One idea of choosing this
hyperparameter in a more principled manner could consist in using just the framework of this
paper: Constructing an ensemble of consensus clusterings and then using a consensus of those
as a hyperprior, which may yield a robust and reasonable choice.

\vspace{3cm}

%\listoftodos
%\acks{}

%---------------------------------------------
%APPENDIX
%---------------------------------------------
\newpage
\appendix

\section{Clustering quality measures} \label{sec:appendix}
This appendix serves as an overview of the used comparison measures in Table
\ref{tab:digits} from Section \ref{sec:digits}.
Throughout this appendix we assume
a clustering $\mathcal{C} = \{C_1, \dots C_k\}$ on a data set $D$
and an external clustering
$\mathcal{C}^* = \{\mathcal{C}^*_1, \dots \mathcal{C}^*_{k^*}\}$. For our experiment,
this external partition is given in terms of the associated consensus.
\paragraph*{Adjusted Rand index.}
The adjusted Rand index (ARI)
is an external measure with values in $[-1,1]$ described in \cite{rand71} and \cite{hubert85}, see also \ref{sec:comparison}.
Based on the values $s_{ij} = |C_i \cap \mathcal{C}^*_j|$,
\begin{align*}
\textnormal{ }
s_i = \sum\limits_{j=1}^{k^*} s_{ij} \textnormal{  and  } s^*_j = \sum\limits_{i=1}^k s_{ij},
\end{align*} it is given as
\begin{align*}
\mathrm{ARI}(\mathcal{C},\mathcal{C}^*) = \frac{ \sum_{ij} \binom{s_{ij}}{2} %
- \big[ \sum_i \binom{s_i}{2} \sum_j \binom{s^*_j}{2} \big] / \binom{|D|}{2} }%
{\frac{1}{2} \big[ \sum_i \binom{s_i}{2} + \sum_j \binom{s^*_j}{2} \big]%
- \big[ \sum_i \binom{s_i}{2} \sum_j \binom{s^*_j}{2} \big] / \binom{|D|}{2} }.
\end{align*}
Larger ARI values indicate partitions with a higher similarity. An ARI value of $1.0$ describes a perfect matching of two
clusterings and under consideration of label permutations.

\paragraph*{Cali\'nski-Harabasz index.} The Cali\'nski-Harabasz index \citep[CHI, see ][]{calinski74}
is an internal measure with values in $[0,\infty)$ defined by
\begin{align*}
\mathrm{CHI}(\mathcal{C}) =
\dfrac{\frac{1}{k-1} \sum\limits_{i=1}^k \Vert \mu_i - \mu \Vert^2}%
{\frac{1}{|D|-k} \sum\limits_{i=1}^k \sum\limits_{x \in C_i} \Vert x - \mu_i \Vert^2 },
\end{align*} where $\mu_i$ is the barycenter of cluster $C_i$
and $\mu$ is the barycenter of the ground set $D$. A higher
value indicates a higher ratio of between-cluster-variance and
within-cluster-variance, which is interpreted as a better overall clustering.

\paragraph*{Dunn-type indices.} The Dunn-type indices \citep[see ][]{dunn74} are internal
metrics which are usually defined as a ratio of intercluster distances and intracluster
diameters. Several different indices can be formulated by this approach. We use the following
definitions:
\begin{align*}
\mathrm{DI_1}(\mathcal{C}) &=
\dfrac{
\min \limits_{i \neq j}
\left\{\inf\limits_{x \in C_i, y \in C_j} \Vert x-y \Vert \right\}}%
{\max\limits_{i} \left\{ \max\limits_{x,y \in C_i} \Vert x-y \Vert \right\}}, \\[0.5cm]
\mathrm{DI_2}(\mathcal{C}) &=
\dfrac{
\frac{1}{\binom{k}{2}}\sum\limits_{i \neq j} \inf\limits_{x \in C_i, y \in C_j} \Vert x-y
\Vert }%
{\frac{1}{k} \sum\limits_{i=1}^n \max\limits_{x,y \in C_i} \Vert x-y \Vert}
\end{align*}
A higher Dunn-index is interpreted as indication of a clustering
that is able to separate the data
better according to (average) distances of observations with respect to their associated
clusters.

\paragraph*{Silhouette coefficient.} The silhouette coefficient is an internal
measure with values in $[-1,1]$. It was first introduced in \cite{rousseeuw87silhouette} with
the form
\begin{align*}
\mathrm{Silhouette}(\mathcal{C})= \dfrac{1}{|D|} \sum\limits_{x \in D} \dfrac{b(x)-a(x)}{\max
\{a(x),b(x)\}},
\end{align*} where we assume $x \in C$ and set
\begin{align*}
a(x) &= \dfrac{1}{|C|} \sum\limits_{y \in C} \Vert x-y \Vert_2 \textnormal{ and}\\
b(x) &= \min\limits_{A \neq C} \bigg( \dfrac{1}{|A|} \sum\limits_{y \in A} \Vert x-y \Vert_2
\bigg).
\end{align*} A higher value of the silhouette coefficient indicates a better average matching
of
all objects in the ground set to their associated cluster. Note that silhouette coefficients
can also be computed for subsets of the clustering or single clusters. However, we only use
the given version of the coefficient on the whole clustering.

%%%%%%%%%%%%%%%%%%%%%%%%%%%%%%%%%%%%%%%%%%%%%%%%%%%%%%%%
%BIBLIOGRAPHY
%%%%%%%%%%%%%%%%%%%%%%%%%%%%%%%%%%%%%%%%%%%%%%%%%%%%%%%%
\bibliographystyle{plainnat}
{\small
\bibliography{references}}

\end{document}